\documentclass[letterpaper]{article}  
\usepackage[a4paper,top=1in,bottom=1in,left=1in,right=1in,marginparwidth=1.75cm]{geometry}
\usepackage{amsfonts}
\usepackage{amsmath}
\usepackage{graphicx}
\usepackage{algorithmic, algorithm}
\usepackage{tabularx}

\newcommand{\normal}{\mathcal{N}}
\newcommand{\eye}{\mathbf{I}}
\newcommand{\zvec}{\mathbf{z}}
\newcommand{\tvec}{\mathbf{t}}
\newcommand{\xvec}{\mathbf{x}}
\newcommand{\yvec}{\mathbf{y}}
\newcommand{\fshared}{f_{\theta_s}}
\newcommand{\ftarget}{f_{\theta_t}}

\newcommand{\data}{\mathcal{D}}
\newcommand{\E}{\mathbb{E}}
\newcommand{\hide}[1]{}
\newcommand{\KL}[2]{\text{KL}(#1 \text{ } || \text{ }#2)}

\newcommand{\elbo}{\mathcal{L}}

\title{Unsupervised Learning with Contrastive Latent Variable Models}
\author{Kristen A. Severson, Soumya Ghosh, and Kenney Ng \\ Center for Computational Health and MIT-IBM Watson AI Lab \\ IBM Research, Cambridge MA}
\date{}

\begin{document}
	\maketitle 
	
	\begin{abstract}
		In unsupervised learning, dimensionality reduction is an important tool for data exploration and visualization. Because these aims are typically open-ended, it can be useful to frame the problem as looking for patterns that are enriched in one dataset relative to another. These pairs of datasets occur commonly, for instance a population of interest vs. control or signal vs. signal free recordings. However, there are few methods that work on \emph{sets} of data as opposed to data points or sequences. Here, we present a probabilistic model for dimensionality reduction to discover signal that is enriched in the \emph{target} dataset relative to the \emph{background} dataset. The data in these sets do not need to be paired or grouped beyond set membership. By using a probabilistic model where some structure is shared amongst the two datasets and some is unique to the target dataset, we are able to recover interesting structure in the latent space of the target dataset. The method also has the advantages of a probabilistic model, namely that it allows for the incorporation of prior information, handles missing data, and can be generalized to different distributional assumptions. We describe several possible variations of the model and demonstrate the application of the technique to de-noising, feature selection, and subgroup discovery settings.
	\end{abstract}
	
	\section{Introduction}
	In unsupervised learning, the goal is often to learn what is unique or interesting about a dataset. Given the subjective nature of this question, it can be useful to frame the problem in the context of what signal is enriched in one dataset, referred to as the \emph{target}, relative to a second dataset, referred to as the \emph{background}. An example of this is an exploration of a heterogeneous disease population, such as patients with Parkinson's disease. The interesting sources of variation are those that are unique to the disease population. However, it is likely that some sources of variation are unrelated to the disease state, for instance variation due to aging. This is difficult to assess without a baseline population, therefore, it is useful to contrast the disease population with a population of healthy controls. Such \emph{contrastive analysis} can discover nuisance variation that is common amongst the two populations and is uninteresting for the problem while highlighting variation unique to the disease population enabling downstream applications such as subgroup discovery.
	
	Despite this natural setting for unsupervised learning, most techniques address individual data points, sequences, or paired data points. Few techniques generalize to the contrastive scenario where we have sets of data but no obvious correspondence between their members. Yet, there are many cases where datasets that can be used in a comparative setting arise naturally: control vs. study populations, pre- and post-intervention groups, and signal vs. signal free groups \cite{Abid2018}. Each of these settings has possible nuisance variation, for example, population level variation, effects unrelated to intervention, and sensor noise variation. 
	
	The recently published contrastive principal component approach (cPCA) \cite{Abid2018} is one example of a technique that can be used for sets of data. cPCA builds on principal component analysis (PCA) \cite{Hotelling1933}, a dimensionality reduction technique which projects data into a lower dimensional space while minimizing the squared loss. PCA and other dimensionality reduction techniques are popular because they allow high-dimensional data to be visualized while removing noise. cPCA seeks to find a projection to a lower dimensional space that discovers variation that is enriched in one dataset as compared to another by applying PCA to the empirical covariance matrix
	\begin{equation}
	C = \frac{1}{n}\sum_{i = 1}^{n}\textbf{x}_i \textbf{x}_i^\textrm{T} - \alpha \frac{1}{m} \sum_{j = 1}^{m} \textbf{y}_j \textbf{y}_j^\textrm{T}
	\label{eq:1}
	\end{equation}
	where $\{\textbf{x}_i\}$ are the observations of interest, $\{\textbf{y}_j\}$ are the comparison data, and $\alpha$ is a tuning parameter. The choice of $\alpha$ is a trade-off between maximizing the retained variance of the target set and minimizing the retained variance of the background set. 
	
	In this work, we develop probabilistic latent variable models applicable to the setting where contrastive analysis is desired. These models are based on the insight that it is possible to emphasize latent structures of interest while suppressing spurious, uninteresting variance in the data through carefully designed statistical models. Such models have several key advantages over deterministic approaches: it is straight forward to incorporate prior domain knowledge, missing and noisy data can naturally be modeled through appropriate noise distributions, model and feature selection can be performed through sparsity promoting prior distributions, and the model can more easily be  incorporated into larger probabilistic systems in a principled manner. Through this paper, we advance the state-of-the-art in several ways. First, we develop latent variable models capable of contrastive analysis. We then demonstrate the generality of our framework by demonstrating how robust and sparse contrastive variants can be developed, learned and how automatic model selection can be performed. We also develop contrastive variants of the variational autoencoder, a deep generative model, and demonstrate its utility in modeling the density of noisy data. Finally, we vet our proposed models through extensive experiments on real world scientific data to demonstrate the utility of the proposed framework.  
	
	\section{Contrastive Latent Variable Models}
	To achieve the aim of discovering patterns that are enriched in one dataset relative to another, we propose a latent variable model where some structure is shared across the two datasets and some structure is unique to the target dataset. Given a target dataset $\{\textbf{x}_i\}_{i=1}^n$ and a background dataset $\{\textbf{y}_j \}_{j=1}^m$, the model is specified
	\begin{equation}
	\begin{gathered}
	\textbf{x}_i = \textbf{Sz}_i + \textbf{W}\textbf{t}_i + \boldsymbol{\mu}_x + \boldsymbol{\epsilon}_i,~~i = 1 \dots n \\
	\textbf{y}_j = \textbf{Sz}_j + \boldsymbol{\mu}_y + \boldsymbol{\epsilon}_j,~~j = 1 \dots m \\
	\end{gathered}
	\label{eq:main}
	\end{equation}
	where $\textbf{x}_i, \textbf{y}_j \in \mathbb{R}^d$ are the observed data, $\textbf{z}_i, \textbf{z}_j \in \mathbb{R}^k$ and $\textbf{t}_i \in \mathbb{R}^t$ are the latent variables, $\textbf{S} \in \mathbb{R}^{d \times k}$ and $\textbf{W} \in \mathbb{R}^{d \times t}$ are the corresponding factor loadings, $\boldsymbol{\mu}_x, \boldsymbol{\mu}_y \in \mathbb{R}^d$ are the dataset-specific means and $\boldsymbol{\epsilon}_i, \boldsymbol{\epsilon}_j\in \mathbb{R}^d$ are the noise. In general, we do not expect the number of samples in the two datasets to be the same, i.e. $n \neq m$. Furthermore, there is no special relationship between the samples $i$ and $j$ in equation~\ref{eq:main}. The primary variables of interest are $\{\textbf{t}_i\}_{i=1}^n$, which are the lower dimensional representation that is unique to the target dataset. 
	
	\subsection{Gaussian likelihood and priors}	
	To provide intuition into why eqn.~\ref{eq:main} meets our goal of capturing patterns enriched in the target with respect to the background, we consider the case where the noise follows isotropic Gaussian distributions, $\boldsymbol\epsilon_i \sim \mathcal{N}(0, \sigma^2 \textbf{I}_d)$ and $\boldsymbol\epsilon_j \sim \mathcal{N}(0, \sigma^2 \textbf{I}_d)$ and the latent variables are modeled using standard Gaussian distributions 	
	\begin{equation}
	\begin{gathered}
	\textbf{x}_i|\textbf{z}_i, \textbf{t}_i \sim \mathcal{N}(\textbf{S}\textbf{z}_i + \textbf{W}\textbf{t}_i + \boldsymbol{\mu}_x, \sigma^2 \textbf{I}_d) \\
	\textbf{y}_j|\textbf{z}_j \sim \mathcal{N}(\textbf{Sz}_j + \boldsymbol{\mu}_y, \sigma^2 \textbf{I}_d) \\
	\textbf{z}_i \sim \mathcal{N}(0,\textbf{I}_k),\quad \zvec_j \sim \mathcal{N}(0,\textbf{I}_k), \quad
	\tvec_i \sim \mathcal{N}(0,\eye_t), \\
	\end{gathered}
	\end{equation}
	where $\mathcal{N}(\mu, \Sigma)$ is a multivariate normal distribution parameterized by mean $\mu$ and covariance $\Sigma$ and $\eye_d$ denotes a $d\times d$ identity matrix. The resulting marginal distributions for the observed data are
	\begin{equation}
	\begin{gathered}
	\xvec_i \sim \mathcal{N}(\boldsymbol{\mu}_x, \textbf{WW}^\textrm{T} + \textbf{SS}^\textrm{T} + \sigma^2\textbf{I}_d) \\
	\yvec_j \sim \mathcal{N}(\boldsymbol{\mu}_y, \textbf{SS}^\textrm{T} + \sigma^2 \textbf{I}_d).		
	\end{gathered}
	\end{equation}
	The covariance structure for the target data is additive and contains a term ($\textbf{SS}^\textrm{T})$ that is shared with the background data and a term that is unique to the target data ($\textbf{WW}^\textrm{T}$). This constructions allows the factor loading $\textbf{W}$ to model the structure unique to the target. The model closely mirrors probabilistic PCA (PPCA) \cite{Tipping1999,Roweis1998} and is exactly PPCA applied to the combined datasets when the target factor loading dimensionality $t$ is zero. Similarly, this model is exactly PPCA applied to only the target dataset when the shared factor loading dimensionality $k$ is zero. Expectation-maximization (EM) \cite{Dempster1977} can be used to solve for the model parameters. Because EM requires conjugacy, most model formulations will not be solved this way. However, we present a summary of the EM steps to provide an intuition about the model. To provide interpretable equations in the below description, we consider the case where the factor loading matrices $\textbf{W}$ and $\textbf{S}$ are orthogonal. 
	
	The model parameters are $\textbf{S}$, $\textbf{W}$, $\mu_x$, $\mu_y$, $\sigma^2$ and the latent variables are $\textbf{z}_i$, $\textbf{z}_j$, $\textbf{t}_i$. The lower bound of the likelihood is 
	\begin{equation}
	\begin{split}
	\mathcal{L} = \sum_{i=1}^{n} \mathbb{E}&_{p(\textbf{z}_i, \tvec_i | \xvec_i)}[\ln p(\textbf{z}_i, \textbf{t}_i, \textbf{x}_{i})] + \\& \sum_{j = 1}^{m} \mathbb{E}_{p(\zvec_j|\yvec_j)}[\ln p(\textbf{z}_j, \textbf{y}_j)]
	\end{split}
	\end{equation}
	The M-step maximizes the lower bound of the likelihood with respect to the parameters. The update step for the shared factor loading is
	\begin{equation}
	\begin{split}
	\tilde{\textbf{S}} = \big[ (\textbf{B} + (\textbf{I} - \textbf{W}&\textbf{R}^{-1} \textbf{W}^\textrm{T})\textbf{T}) \textbf{S} \big] \\ &(\sigma^2\textbf{I} + \textbf{M}^{-1}\textbf{S}^\textrm{T}(\textbf{B} + \textbf{T})\textbf{S})^{-1}
	\end{split}
	\label{eq:16}
	\end{equation}
	where $\textbf{B}$ is the sample covariance of the background data, $\textbf{T}$ is the sample covariance of the target data, $\textbf{M} = \sigma^2 \textbf{I}_k + \textbf{S}^\textrm{T}\textbf{S}$, and $\textbf{R} = \sigma^2 \textbf{I}_t + \textbf{W}^\textrm{T}\textbf{W}$. 
	The update step for the target factor loading is
	\begin{equation}
	\tilde{\textbf{W}} = ((\textbf{I} - \textbf{SM}^{-1}\textbf{S})\textbf{TW})(\sigma^2\textbf{I} + \textbf{R}^{-1}\textbf{W}^\textrm{T}\textbf{TW})^{-1}
	\label{eq:18}
	\end{equation}
	Details on the derivation can be found in the supplemental information. It is useful to recall that the orthogonal projection onto the range space of a matrix $\textbf{A}$ is given by $\textbf{P} = \textbf{A}(\textbf{A}^\textrm{T}\textbf{A})^{-1}\textbf{A}^\textrm{T}$ and the orthogonal projection onto the nullspace of $\textbf{A}$ is given by $\textbf{I} - \textbf{P}$. In eqn.~\ref{eq:18}, $\textbf{I} - \textbf{SM}^{-1}\textbf{S}$ can be expanded using the definition of $\textbf{M}$ to $\textbf{I} - \textbf{S}(\sigma^2\textbf{I}_k + \textbf{S}^\textrm{T}\textbf{S})^{-1}\textbf{S}^\textrm{T}$. Similarly, in eqn.~\ref{eq:16}, $\textbf{I} - \textbf{W} \textbf{Q}^{-1}\textbf{W}^\textrm{T}$ can be expanded using the definition of $\textbf{Q}$ to $\textbf{I} - \textbf{W}(\sigma^2 \textbf{I}_t + \textbf{W}^\textrm{T}\textbf{W})^{-1}\textbf{W}^\textrm{T}$. When $\sigma^2$ is small, these equations are similar to the projection onto the nullspace of $\textbf{S}$ and $\textbf{W}$, respectively. This matches our intuition as to how these factor loading matrices are updated: in a sense, the part of the target data captured by the target factor loading space is \emph{projected away} before updating the shared factor loading space, and vice versa. This behavior is similar to cPCA. In eqn.~\ref{eq:1}, as $\alpha$ goes to infinity, directions not in the null space of the background data covariance are given an infinite penalty. When this is the case, cPCA projects the target data onto the null space of the background data and then performs PCA \cite{Abid2018}. 
	
	The update steps can also be compared to the PPCA updates. For the factor loading matrix $\textbf{W}$, the update step is:
	\begin{equation}
	\tilde{\textbf{W}} = \textbf{T}\textbf{W}(\sigma^2\textbf{I} + \textbf{R}^{-1}\textbf{W}^\textrm{T}\textbf{T}\textbf{W})^{-1}
	\end{equation}
	which is the same as eqn.~\ref{eq:18}, except for the projection term.
	
	\subsection{Beyond Gaussian Models}
	The assumptions of Gaussianity are not necessary for recovering latent structure that enriches desired patterns in the target dataset. We can more generally express the proposed model as: 
	\begin{equation}
	\small{
		\begin{split}
		p(\data, &\{\zvec_i, \tvec_i\}_{i=1}^n, \{\zvec_j \}_{j=1}^m; \Theta) = \\
		&p(\Theta)\prod_{i=1}^n p(\textbf{x}_i| \textbf{z}_i, \textbf{t}_i; \textbf{W}, \textbf{S}, \boldsymbol{\mu}_x, \sigma^2) p(\zvec_i)p(\tvec_i)\\ &\prod_{j=1}^m p( \yvec_j|  \zvec_j; \textbf{S},\boldsymbol{\mu}_y, \sigma^2) p(\zvec_j),
		\end{split}
	}
	\label{eq:baseModel}
	\end{equation}
	where $\data = \{\{\textbf{x}_i\}_{i=1}^n, \{\textbf{y}_j \}_{j=1}^m\}$ and $\Theta = \{ \textbf{W}, \textbf{S}, \boldsymbol{\mu}_x, \boldsymbol{\mu}_y, \sigma^2 \}$. The primary modeling decisions are to choose the appropriate likelihoods and priors on the loading matrices. The particular choices are governed by the application and domain specific knowledge.
	
	However, this flexibility comes at a price: the posterior distributions $p(\tvec_i, \zvec_i, \zvec_j | \data)$ are no longer guaranteed to be tractable. Consequently, the EM algorithm sketched in the previous section is no longer available and instead, we use variational inference~\cite{wainwright2008graphical}. In summary, the intractable posteriors are approximated with tractable surrogates $q(\tvec_i | \lambda_{\tvec_i})q(\zvec_i | \lambda_{\zvec_i})q(\zvec_j | \lambda_{\zvec_j})$ and divergence $\KL{q}{p}$ is minimized with respect to the variational parameters $\lambda = \{\{\lambda_{\zvec_i}, \lambda_{\tvec_i}\}_{i=1}^n, \{\lambda_{\zvec_j}\}_{j=1}^m\}$. This is equivalent to maximizing the lower bound of the marginal likelihood,
	\begin{equation}
	\small{
		\begin{split}
		p(\data; \Theta) &\geq \mathcal{L}(\lambda, \Theta)\\
		= &\sum_i \E_{q(\zvec_i; \lambda_{\zvec_i})q(\tvec_i; \lambda_{\tvec_i})}[\text{ln } p(\xvec_i | \zvec_i, \tvec_i; \Theta_{ \setminus \{\boldsymbol{\mu}_y\}})]\\
		&- \KL{q(z_i ; \lambda_{\zvec_i})}{p(\zvec_i)} - \KL{q(\tvec_i;\lambda_{\tvec_i})}{p(\tvec_i)} \\
		&+ \sum_j \E_{q(\zvec_j; \lambda_{\zvec_j})}[\text{ln } p(\yvec_j | \zvec_j; \Theta_{ \setminus \{\boldsymbol{\mu}_x, \theta_x\}})] \\
		&- \KL{q(\zvec_j ; \lambda_{\zvec_j})}{p(\zvec_j)} + \text{ln }p(\Theta)
		\end{split}
	}
	\label{eq:vi}
	\end{equation} 
	where $\Theta_{ \setminus \{\cdot \}}$ implies the parameters in $\Theta$ except the parameters denoted in the set. Depending on the choice of $q$ and $p$ the expectations required for computing $\elbo(\lambda, \Theta)$ may themselves be intractable. We use recently proposed black box techniques~\cite{Ranganath2014,Kingma2014,Rezende2014,Titsias2014} to sidestep this additional complication. In particular, we approximate the intractable expectations in $\elbo(\Theta, \lambda)$ with unbiased Monte-Carlo estimates, $\tilde{\elbo}(\Theta, \lambda)$. Because the latent variables of interest are continuous, we are able to use reparameterization gradients~\cite{Kingma2014,Rezende2014} to differentiate through the sampling process and obtain low variance estimates of $\nabla_{\lambda,\Theta}\mathcal{L}(\Theta, \lambda)$, $\nabla_{\lambda,\Theta}\tilde{\elbo}(\Theta, \lambda)$. Using the noisy but unbiased gradients, optimization can proceed using a stochastic gradient ascent variant, e.g. ADAM~\cite{kingma2014adam}. In our experiments we use Edward~\cite{Tran2016}, a library for probabilistic modeling, to implement these inference strategies for the proposed models. We sketch the pseudocode for variational learning in Algorithm~\ref{alg:overall}.
	\begin{algorithm}[ht]
		\caption{Pseudocode}
		\label{alg:overall}
		\begin{algorithmic}[1]
			\STATE \textbf{Input} Model $p(\data; \Theta)$, variational approximations  $q(\{z_i, t_i\}_{i=1}^n, \{z_j\}_{j=1}^m\mid\lambda)$
			\STATE \textbf{Output}: Optimized $\Theta$ and variational parameters $\lambda$  
			\STATE Initialize $\lambda$ and $\Theta$.
			\REPEAT
			\STATE Use reparameterization trick to compute unbiased estimates of the gradients of the objective in Eqn.~\ref{eq:vi}, $\nabla_{\lambda, \Theta} \tilde{\elbo}(\lambda, \Theta)$ \\
			\STATE Update $\lambda^{(l+1)} \leftarrow \text{ADAM}(\lambda^{(l)}, \nabla_{\lambda} \tilde{\elbo}(\lambda, \Theta))$,  $\Theta^{(l+1)} \leftarrow \text{ADAM}(\Theta^{(l)}, \nabla_{\Theta} \tilde{\elbo}(\lambda, \Theta))$
			\UNTIL {convergence}
		\end{algorithmic}
	\end{algorithm}
	
	Finally, we note that the black box inference framework does not restrict us to point estimates of $\Theta$. As we will illustrate in the next section, it is possible to infer variational distributions over $\Theta$ by specifying an appropriate approximation $q(\Theta \mid \lambda_\Theta)$.

	\subsection{cLVM Variants }
	We refer to the base structure of the model as provided in eqn.~\ref{eq:baseModel} as a contrastive latent variable model, cLVM. As previously noted, different choices for the distributions in eqn.~\ref{eq:baseModel} can be made to address the specific challenges of the application. Several models are introduced here and are summarized in Table~\ref{tab:modelVar}.
	
	\begin{table*}[]
		\centering
		\begin{tabularx}{\textwidth}{@{\extracolsep{\fill} }lccc}
			Model name & Prior & Likelihood & Variational Approximation \\ \hline\noalign{\smallskip}
			cLVM  &  --  & Gaussian & --  \\ \hline\noalign{\smallskip}
			Sparse cLVM & \begin{tabular}[r]{@{}r@{}} 
				
				$p(\textbf{W}) = \prod_{i = 1}^{d} \mathcal{N}(\textbf{W}_{i:} | \rho_i, \tau)$ \\ $C^+(\rho_i |0, 1)C^+(\tau | 0, b_g) $         
			\end{tabular}              & Gaussian                         & \begin{tabular}[c]{@{}c@{}} 
				$q(\textbf{W}) = \mathcal{N}(\cdot, \cdot)$ \\ $q(\ln \boldsymbol{\rho}) = \mathcal{N}(\cdot, \cdot)   $  \\ $q(\ln \tau) = \mathcal{N}(\cdot, \cdot)$
			\end{tabular}                           \\ \hline\noalign{\smallskip}
			\begin{tabular}[l]{@{}l@{}} 
				cLVM with \\ model selection  \end{tabular} & $p(\textbf{S}) = \prod_{i = 1}^{d-1} \mathcal{N}(\textbf{S}_{:j}| 0, \alpha_j) \textrm{IG}(\alpha_j | a, b) $   & Gaussian                            & \begin{tabular}[c]{@{}c@{}} 
				$q(\textbf{S}) = \mathcal{N}(\cdot, \cdot)$ \\ $q(\ln \boldsymbol{\alpha}) = \mathcal{N}(\cdot, \cdot)   $ 
			\end{tabular}              \\ \hline\noalign{\smallskip}
			Robust cLVM & $p(\sigma^2) = \textrm{IG}(a, b) $ & Student's t & $q(\ln \sigma^2)= \mathcal{N}(\cdot, \cdot)$ \\ \hline\noalign{\smallskip}
			cVAE & --  & \begin{tabular}[c]{@{}c@{}} 
				Gaussian parameterized \\ by neural network
			\end{tabular}     
			& $q(\zvec_i, \tvec_i) = \mathcal{N}(g_\mu(\cdot), g_\sigma(\cdot))$  \\ \hline
		\end{tabularx}
		\caption{\textbf{Summary of the model variants.} For all of the models in the table, the latent variables $\{\textbf{z}_i, \textbf{t}_i\}_{i=1}^n, \{\textbf{z}_j\}_{j=1}^m$ are modeled as standard Gaussians and the variational distributions are also Gaussian, unless otherwise noted. The model choice depends on the application. The various models are not mutually exclusive and may also be combined.}
		\label{tab:modelVar}
	\end{table*}
	
	\paragraph{Sparse cLVM}	
	One application-specific problem is feature selection. In unsupervised learning, there is often a secondary goal of learning a subset of measurements that are of interest which is motivated by improved interpretability. This is especially important when the observed data is very high-dimensional. For instance, many biological assays result in datasets that have tens of thousands of measurements such as SNP and RNA-Seq data. During data exploration, discovering a subset of these measurements that is important to the target population can help guide further analysis. To learn a latent representation that is only a function of a subset of the observed dimensions, certain rows of the target factor loading, $\textbf{W}$, must be zero. The observed data corresponding to the zero rows in $\textbf{W}$ then have no contribution to the latent representation $\textbf{t}$. Because there is no restriction on $\textbf{S}$, a sparsity requirement for $\textbf{W}$ does not imply that the corresponding observation is zero.
	
	One way to achieve this behavior is by using a regularization penalty on the model parameters. The penalty is added to the objective function to incite certain behavior. Regularization penalties can be related to priors by noting that $\log p(\textbf{W}) \propto r(\textbf{W})$, where $r(\cdot)$ is the penalty function. For feature selection, a group sparsity penalty \cite{Yuan2007} could be used. The rows of $\textbf{W} \in \mathbb{R}^{d \times t}$ are penalized:
	\begin{equation}
	r(\textbf{W}) = \rho \sum_{i = 1}^{d} \sqrt{p_i}\|\textbf{W}_{i:}\|_2
	\end{equation}
	where $\textbf{W}_{i:}$ is the $i^{th}$ row of $\textbf{W}$. This functional form is known to lead to sparsity at the group level, i.e. all members of a group are zero or non-zero. For increasing values of $\rho$, the target factor loading matrix has a larger number of zero-valued rows.
	
	Sparsity inducing priors such as the automatic relevance determination (ARD) \cite{Bishop1999,Virtanen2011,Klami2013} or global-local shrinkage priors such as the horseshoe \cite{Carvalho2009,Carvalho2010} can also be easily incorporated into the framework\. Using the horseshoe prior as an example, the $i^{th}$ row of $\textbf{W}$ is modeled,
	\begin{equation}
	\begin{gathered}
	\textbf{W}_{i:}  | \rho_i, \tau \sim \mathcal{N}(0, \rho_i^2 \tau^2\textbf{I}_t) \\
	\rho_i \sim C^+(0,1),  \quad	\tau \sim C^+(0,b_g)
	\end{gathered}
	\end{equation}
	where $a \sim C^+(0, b)$ is the half-Cauchy distribution with density $p(a|b) = \frac{2}{\pi}b(1 + \frac{a^2}{b^2})$ for $a > 0$. The horseshoe prior is useful for subset selection because it has heavy tails and an infinite spike at zero. Further discussion can be found in the supplemental information. For both the prior and regularization formulations, \emph{groups} of rows in $\textbf{W}$ could also be used instead of single rows if such a grouping exists.
	
	
	\paragraph{cLVM with Automatic Model Selection}		
	The ARD prior is more typically applied to the columns of a factor loading matrix. This use allows for automatic selection of the dimension of the matrix. This could also be done in the cLVM model. Although both latent spaces can have any dimension less than $d$, which must be selected, we generally recommend setting the target dimension to two for visualization purposes. To select the dimension of the shared space, the percent variance explained can be analyzed or a prior, such as the ARD prior can be used. The columns of $\textbf{S}$ are modeled
	\begin{equation}
	\textbf{S}_{:j} | \alpha_j \sim \mathcal{N}(0, \alpha_j \textbf{I}_d), \quad 
	\alpha_j \sim \text{IG}(a_0, b_0).
	\end{equation}
	The ARD prior has been shown to be effective at model selection for PPCA models \cite{bishop1999bayesian}.
	
	\paragraph{Robust cLVM} Another application-specific goal may be to systematically handle outliers in the dataset. Similar to PPCA, the cLVM model is sensitive to outliers and can produce poor results if outliers are not addressed. It may be possible to remove outliers from the dataset, however this is typically a manual process that requires domain expertise and an understanding of the process that generated the data. A more general approach to handling outliers uses a heavy-tailed distribution to describe the data. One approach for constructing heavy tailed distributions is through scale mixtures of Gaussians~\cite{west1987scale}. Consider,  
	\begin{equation}
	\sigma^2 \sim \textrm{IG}(a,b).
	\end{equation}
	The resulting marginal distribution of the observed data is
	\begin{equation}
	\begin{split}
	p(\textbf{x}_i | \mu, a, b) &=  \prod_{k = 1}^{d} \int_{0}^{\infty} \mathcal{N}(\textbf{x}_{ik} | \boldsymbol{\mu}, \sigma^2) \textrm{IG}(\sigma^2 | a, b) d\sigma^2 \\
	&= \prod_{k = 1}^{d} \textrm{St}(\textbf{x}_{ik} |\boldsymbol{\mu}, \nu = 2a, \lambda = \frac{a}{b})
	\end{split}
	\end{equation}
	where $\textrm{St}$ indicates a Student's t-distribution \cite{Archambeau2006}. The larger probability mass in the tails of the Student's t-distribution, as compared to the normal distribution, allows the model to be more robust to outliers. 
	
	\subsection{Beyond Linear Models}
	\paragraph{Contrastive Variational Autoencoders}\label{sec:vae}
	Thus far we have only considered models that linearly map latent variables $\zvec$ and $\tvec$ to the observed space. The linearity constraint can be relaxed, and doing so leads to powerful generative models capable of accounting for nuisance variance.
	\begin{equation}
	\begin{gathered}
	\xvec_i = \fshared(\zvec_i) + \ftarget(\tvec_i) + \boldsymbol\epsilon_i, \quad i=1 \ldots n \\
	\yvec_j = \fshared(\zvec_j) + \boldsymbol\epsilon_j, \quad j=1 \ldots m,
	\end{gathered}
	\end{equation}
	where $\boldsymbol\epsilon_i \sim \mathcal{N}(0, \sigma^2)$ , $\boldsymbol\epsilon_j \sim \mathcal{N}(0, \sigma^2)$, and $\fshared$, $\ftarget$ represent non-linear transformations parameterized by neural networks. The latent variables are modeled using standard Gaussian distributions, as before. Observe that similar to the linear case (eqn.~\ref{eq:main}) the target and background data share the projection $\fshared$ while the target retains a private projection $\ftarget$. This construction forces $\fshared$ to model commonalities between the target and background data while allowing $\ftarget$ to capture structure unique to the target. 
	
	This model can be learned by maximizing the lower bound to the marginal likelihood $p(\data | \Theta)$, $\Theta = \{ \theta_s, \theta_t, \mu_x, \mu_y, \sigma^2 \}$, analogously to eqn.~\ref{eq:vi}. However, a large amount of data is typically required to learn such a non-linear model well. Moreover, since the number of latent variables proliferate with increasing data, it is computationally more efficient to amortize the cost of inferring the latent variables through inference networks shared between the data instances. In particular, we parametrize the variational posteriors $q_{\lambda_t}(\zvec_i, \tvec_i | \xvec_i) = \normal(\zvec_i | g^\mu_{\lambda_t}(\xvec_i), g^\sigma_{\lambda_t}(\xvec_i)) \normal(\tvec_i | g^\mu_{\lambda_t}(\xvec_i), g^\sigma_{\lambda_t}(\xvec_i)$ and $q_{; \lambda_s}(\zvec_j | \yvec_j) = \normal(\zvec_j | g^\mu_{\lambda_s}(\yvec_j), g^\sigma_{\lambda_s}(\yvec_j))$, where $\lambda_t$ and $\lambda_s$ are inference network parameters. Unlike eqn.~\ref{eq:vi} where the variational parameters grow with the number of data instances, the variational parameters $\lambda_t$ and $\lambda_s$ do not. $\lambda_t$ is shared amongst the target instances while $\lambda_s$ is shared between the background examples. This is an example of \emph{amortized variational inference}~\cite{dayan1995helmholtz,gershman2014amortized}. Finally, learning proceeds by maximizing the evidence lower bound,
	\begin{equation}
	\small{
		\begin{aligned}[b]
		& p(\data; \Theta) \geq \elbo(\Theta, \lambda_s, \lambda_t) \\
		& = \sum_i \E_{q_{\lambda_t}(\zvec_i, \tvec_i | \xvec_i)}[\text{ln } p(\xvec_i | \zvec_i, \tvec_i; \Theta_{ \setminus \{\boldsymbol{\mu}_y\}})]\\
		&- \KL{q_{\lambda_t}(\zvec_i, \tvec_i | \xvec_i)}{p(\zvec_i)p(\tvec_i)} \\
		&+ \sum_j \E_{q_{\lambda_s}(\zvec_j | \yvec_j)}[\text{ln } p(\yvec_j | \zvec_j; \Theta_{ \setminus \{\boldsymbol{\mu}_x, \theta_x\}})] \\
		&- \KL{q_{\lambda_s}(\zvec_j | \yvec_j)}{p(\zvec_j)} + \text{ln }p(\Theta),
		\label{eq:vae}
		\end{aligned}
	}
	\end{equation}
	with respect to $\Theta$ and $\lambda_s$, $\lambda_t$. The KL terms are available to us in closed form, however the expectation terms are intractable and we again resort to Monte Carlo approximations and re-parameterized gradients to enable stochastic gradient ascent. We refer to this combination of the non-linear model and the amortized variational inference scheme as the contrastive variational auto encoder (cVAE).
	
	\section{Related Work}
	There are many techniques for dimensionality reduction, e.g. \cite{Hotelling1933,Maaten2008,Cox2008}. This review focuses on dimensionality techniques that use sets of data and/or address issues related to nuisance variation. Canonical correlation analysis (CCA) \cite{Hotelling1936} and its probabilistic variant (PCCA) \cite{Bach2005} use two (or more) sets of data, however requires that samples are paired views (or higher dimensional sets of views) of the same sample. For instance perhaps several tests are run on a single patient and therefore the tests are linked via the patient identity. In CCA, the number of samples in the sets must be equal, $n=m$, however the dimensionality of each sample does not need to be the same. Damianou, Lawrence and Ek \cite{Damianou2016} proposed a nonlinear extension of PCCA where the mappings are sampled from a Gaussian process. The resulting model is a multi-view extension of GP-LVM \cite{Lawrence2005}, but still requires linking the samples across datasets.
	
	In this work, we propose addressing nuisance variation in the dataset by introducing a structure to the latent representation. Schulam and Saria \cite{Schulam2015} investigate a similar idea with respect to sharing representations across different parts of the full data. In their work, a hierarchical model for disease trajectory is proposed where some of the model coefficients are shared across subsets of the data, e.g. total population and individual. This idea has also been proposed for the unsupervised analysis of time series data \cite{Hsu2017,Li2018}. Data samples are assumed to have a latent representation that can be partitioned into static and dynamic contributions. None of these works have considered a contrastive setting. There has also been work in addressing explicit sources of nuisance variation. Louizos \emph{et al.} \cite{Louizos2016} explores a setting where certain variables within the dataset are a priori identified as nuisance and the remaining variables contribute to the latent representation. The observed data is modeled $\textbf{x} \sim p_\theta(\textbf{z}, \textbf{s})$ where $\textbf{s}$ are the observed nuisance variables. 
	
	\section{Experiments}
	Contrastive latent variable models have applications in subgroup discovery, feature selection, and de-noising, each of which is demonstrated here leveraging different modeling choices. We use examples from \cite{Abid2018} to highlight the similarities and differences between the two approaches. The results of cLVM as applied to synthetic datasets can be found in the supplemental information.
	
	\begin{figure*}[t]
		\centering
		\includegraphics[width=\textwidth]{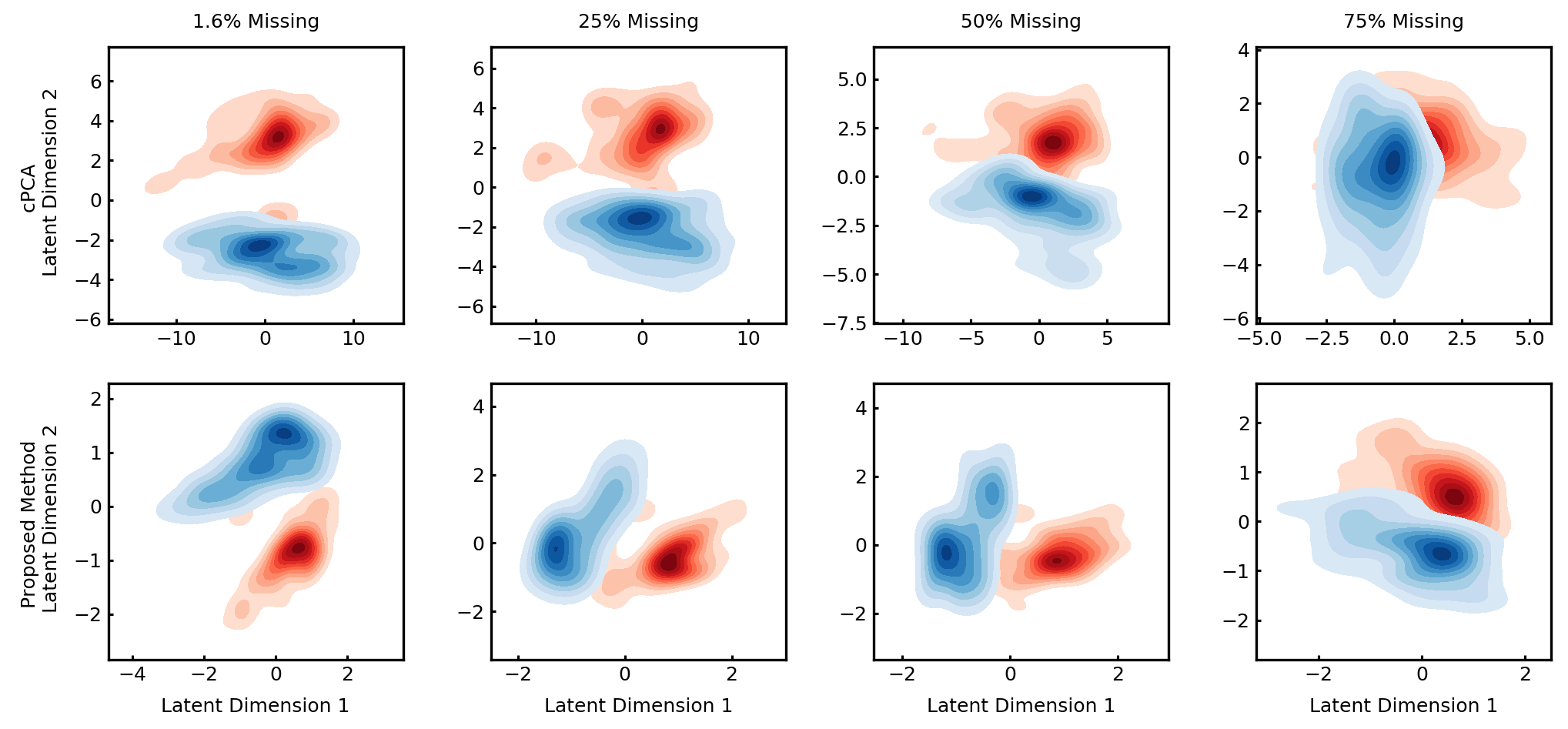}
		\caption{\textbf{cLVM is robust to missing data.} Density plots of the subgroups revealed in the target latent representation of the mice protein expression data. Red and blue points are the control and trisomic mice samples, respectively. The rows use cPCA and robust cLVM to learn the latent representation, respectively. Each column uses a different level of missing data, starting with the leftmost column containing the natural level of missing data. PCA is unable to perform subgroup discovery (see supplemental information) and robust cLVM is better able to perform subgroup discovery in the presence of missing data.}
		\label{fig:micePCCA}
	\end{figure*}
	
	\begin{figure*}
		\centering
		\includegraphics[width=\textwidth]{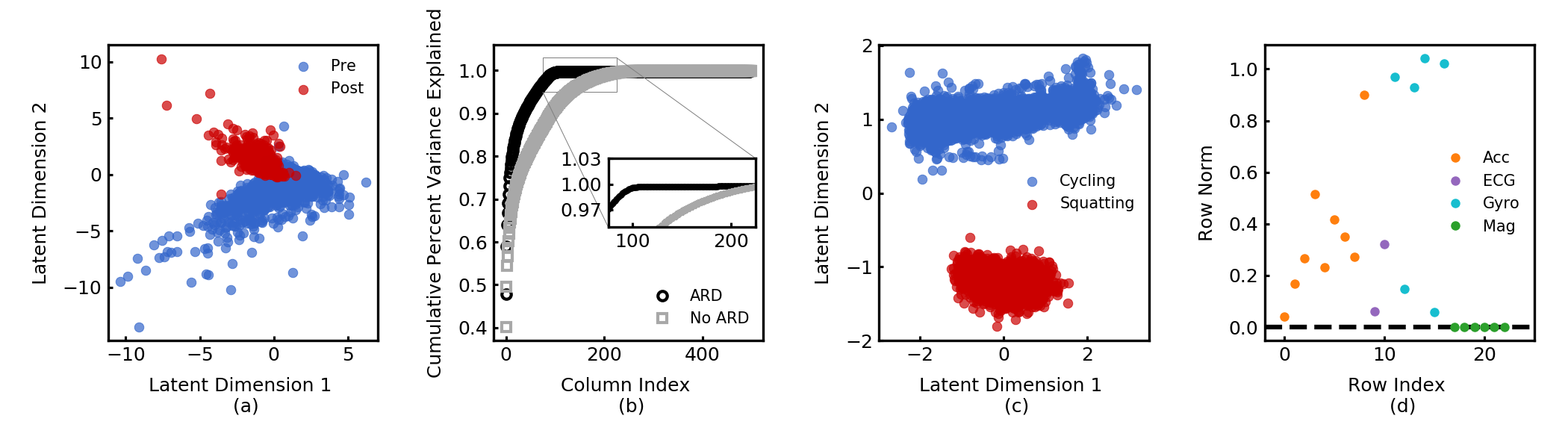}
		\caption{\textbf{cLVM variants allow for model and feature selection.} (a) Subgroups revealed in the target latent representation for the RNA-Seq dataset using the model selection cLVM variant. (b) The percent variance explained by the ordered columns of the shared factor loading for LVM with and without ARD (model selection). The ARD model has over 100 fewer non-zero columns in the shared factor loading. (c) Subgroups revealed in the target latent representation for the mHealth dataset using sparse cLVM. (d) The norms of the rows of the target factor loading for sparse LVM where the different colors correspond to different sensor types. The six dimensions with zero-valued norms correspond to magnetometer readings.}
		\label{fig:RNASeq}
	\end{figure*}
	
	\begin{figure*}[!h]
		\centering
		\includegraphics[width=\textwidth]{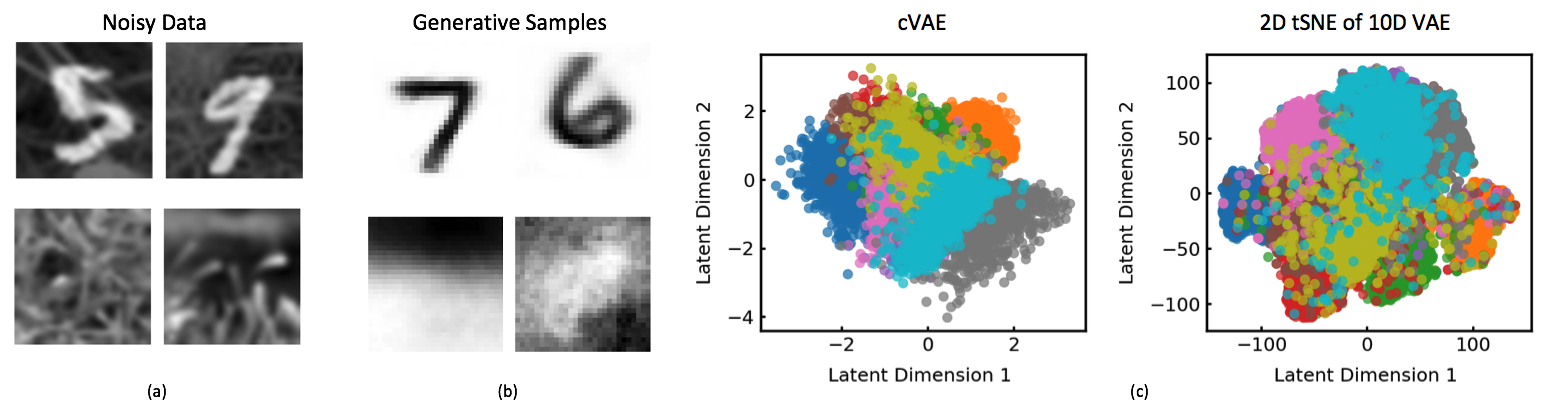}
		\caption{\textbf{cVAE recovers meaningful structure from noisy data.} (a) Samples of the target noisy images of digits on grass and background grass images. (b) Generative samples of the de-noised target (top row) and background (bottom row) which are enabled by the cVAE structure. Note there is no correspondence between the samples in (a) and (b). (c) The 2D cVAE projection and a 2D tSNE projection of a VAE with 10 dimensional space. The colors represent different digits. }
		\label{fig:MNIST_result_horiz}
	\end{figure*}

	\subsection{Subgroup Discovery for Incomplete Data}
	To demonstrate the use of cLVM for subgroup discovery, we use a dataset of mice protein expression levels \cite{Higuera2015}. The target dataset has 270 samples of two unknown classes of mice: trisomic (Down Syndrome model) and control. The background dataset has 135 known control samples. Each sample has 77 measurements. The dataset contains missing values at a level of approximately 1.6\% due to technical artifacts and sampling that cannot be repeated. One of the advantages of the probabilistic approach is that it naturally handles missing data. Depending why the data is missing, missing data can either be ignored, marginalized, or explicitly modeled. For the mice protein dataset, we marginalize over the missing values by treating the missing data as latent variables and adding a corresponding normal variational approximation. Increasing levels of missing data were tested by artificially removing data from the target dataset. The robust variation of the model is applied to account for other possible data issues. The target and shared dimensionalities are both set to two. Fig.~\ref{fig:micePCCA} shows the latent representation using cPCA and robust cLVM for the naturally occurring missing level, 25\%, 50\%, and 75\% missing data. cPCA does not have natural handling for missing data therefore mean imputation was used to first fill-in. PCA is unable to recover the structure in the dataset (see supplemental information for results). Both cPCA and robust cLVM find the subsets, however, the proposed method is better able to discover the subgroups as the amount of missing data increases.

	\subsection{Subgroup Discovery for High Dimensional Data}
	To highlight the use of cLVM for subgroup discovery in high-dimensional data, we use a dataset of single cell RNA-Seq measurements \cite{Zheng2017}. The target dataset consists of expression levels from 7,898 samples of bone marrow mononuclear cells before and after stem cell transplant from a leukemia patient. The background contains expression levels from 1,985 samples from a healthy individual. Pre-processing of the data reduces the dimensionality from 32,738 to 500 \cite{Zheng2017,Abid2018}. Given the size of the data to explore, it is useful in this setting to use an ARD prior to automatically select the dimensionality of the shared latent space. The target latent space is set to two and an $\textrm{IG}(10^{-3}, 10^{-3})$ prior is used for the columns of the shared factor loading. Fig.~\ref{fig:RNASeq}a shows the resulting latent representation, which is able to discover the subgroups, whereas PCA is not (see supplemental information). Fig.~\ref{fig:RNASeq}b compares the percent of variance explained in the ranked columns as compared to the cLVM model without model selection. The model with ARD uses over 100 fewer columns in the shared factor loading matrix and avoids an analysis to manually select the dimension.
	
	
	\subsection{Automatic Feature Selection using Sparse cLVM}
	The third example uses a dataset, referred to as mHealth, that contains 23 measurements of body motion and vital signs from four types of signals \cite{Banos2014,Banos2015}. The participants in the study complete a variety of activities. The target data is composed of the unknown classes of cycling and squatting and the background data is composed of the subjects lying still. In this application, we demonstrate feature selection by learning a latent representation that both separates the two activities and uses only a subset of the signals. A group sparsity penalty is used, as described in the methodology, on the target factor loading. The target dimension is two, the shared dimension is twenty, and $\rho$ is 400. $\rho$ is selected by varying its value and inspecting the latent representation. The latent representation using regularization is shown in Fig.~\ref{fig:RNASeq}c. The two classes are clearly separated. Fig.~\ref{fig:RNASeq}d shows the row-wise norms of the target factor loading. The last six dimensions, corresponding to the magnetometer readings, are all zero which indicates that the magnetometer measurements are not important for differentiating the two classes and can be excluded from further analysis.
	
	\subsection{De-noised Generative Modeling using cVAE}
	Finally, to demonstrate the utility of cVAE, we consider a dataset of corrupted images (see Fig.~\ref{fig:MNIST_result_horiz}a). This dataset was created by overlaying a randomly selected set of $30,000$ MNIST~\cite{LeCun1998} digits on randomly selected images of the grass category from Imagenet~\cite{Russakovsky2015}. The background is $30,000$ grass images. We train a cVAE with a two-dimensional target latent space and an eight-dimensional shared space. We use fully connected encoder and decoder networks with two hidden layers with $128$ and $256$ hidden units employing rectified-linear non-linearities. For the cVAE, both the target and shared decoders $\theta_s$ and $\theta_t$ use identical architectures. We compare against a standard variational autoencoder with an identical architecture and employ a latent dimensionality of ten, to match the combined dimensionality of the shared and target spaces of the contrastive variant. Fig.~\ref{fig:MNIST_result_horiz}c presents the results of this experiment. The latent projections for the cVAE cluster according to the digit labels. VAE on the other hand confounds the digits with the background and fails to recover meaningful latent projections. Moreover, cVAE allows us to selectively generate samples from the target or the background space, Fig.~\ref{fig:MNIST_result_horiz}b. The samples from the target space capture the digits, while the background samples capture the coarse texture seen in the grass images. Additional comparisons with a VAE using a two dimensional latent space is available in the supplemental. 
	\section{Conclusions}
	Dimensionality reduction methods are important tools for unsupervised data exploration and visualization. We propose a probabilistic model for improved visualization when the goal is to learn structure in one dataset that is enriched as compared to another. The latent variable model's core characteristic is that it shares some structure across the two datasets and maintains unique structure for the dataset of interest. The resulting cLVM model is demonstrated using robust, sparse, and nonlinear variations. The method is well-suited to scenarios where there is a control dataset, which is common in scientific and industrial applications.

\bibliographystyle{IEEEtran}
\bibliography{PcPCAbib2}
\newpage
\appendix
\section{Expectation-maximization}

For the EM algorithm, we consider the conditional joint distribution of $\textbf{t}_i$ and $\textbf{z}_i$
\begin{equation}
\begin{split}
p \bigg(\begin{bmatrix}
\textbf{t}_i \\
\textbf{z}_i
\end{bmatrix} \bigg| \textbf{x}_i \bigg) = \mathcal{N}\bigg( \begin{bmatrix}
\textbf{W}^\textrm{T} \textbf{W} + \sigma^2\textbf{I}_t & \textbf{W}^\textrm{T}\textbf{S} \\
\textbf{S}^\textrm{T}\textbf{W} & \textbf{S}^\textrm{T} \textbf{S} + \sigma^2\textbf{I}_k
\end{bmatrix}^{-1} \begin{bmatrix} 
\textbf{W}^\textrm{T} \\
\textbf{S}^\textrm{T}
\end{bmatrix}(\textbf{x}_i - \mu_x), &\\ \sigma^2\begin{bmatrix}
\textbf{W}^\textrm{T} \textbf{W} + \sigma^2\textbf{I}_t & \textbf{W}^\textrm{T}\textbf{S} \\
\textbf{S}^\textrm{T}\textbf{W} & \textbf{S}^\textrm{T} \textbf{S} + \sigma^2\textbf{I}_k
\end{bmatrix}^{-1}
\bigg)
\end{split}
\end{equation}
To provide intuition about the model, we consider the simplifying case where the factor loading matrices are orthogonal, therefore $\textbf{W}^\textrm{T}\textbf{S} = 0$. The distributions can then be written:
\begin{equation}
\begin{gathered}
p(\textbf{z}_i| \textbf{x}_i)	 = \mathcal{N}\bigg((\sigma^2 \textbf{I}_k + \textbf{S}^\textrm{T}\textbf{S})^{-1}\textbf{S}^\textrm{T}(\textbf{x}_i - \mu_x),( \textbf{I}_k + \frac{1}{\sigma^2}\textbf{S}^\textrm{T}\textbf{S})^{-1}  \bigg) \\
p(\textbf{z}_j| \textbf{y}_j)	 = \mathcal{N}\bigg((\sigma^2 \textbf{I}_k + \textbf{S}^\textrm{T}\textbf{S})^{-1}\textbf{S}^\textrm{T}(\textbf{y}_j - \mu_y),( \textbf{I}_k + \frac{1}{\sigma^2}\textbf{S}^\textrm{T}\textbf{S})^{-1}  \bigg) \\
p(\textbf{t}_i| \textbf{x}_i)	 = \mathcal{N}\bigg((\sigma^2 \textbf{I}_k + \textbf{W}_q^\textrm{T}\textbf{W})^{-1}\textbf{W}^\textrm{T}(\textbf{x}_i - \mu_x),( \textbf{I}_k + \frac{1}{\sigma^2}\textbf{W}^\textrm{T}\textbf{W})^{-1}  \bigg).
\end{gathered}
\end{equation}
and the corresponding expectations are:
\begin{equation}
\begin{gathered}
\mathbb{E}_{p(\textbf{z}_i | \textbf{x}_i)} [\textbf{z}_i] = \textbf{M}^{-1}\textbf{S}^\textrm{T}(\textbf{x}_i - \mu_x) \\
\mathbb{E}_{p(\textbf{z}_i | \textbf{y}_i)} [\textbf{z}_j] = \textbf{M}^{-1}\textbf{S}^\textrm{T}(\textbf{y}_j - \mu_y) \\
\mathbb{E}_{p(\textbf{z}_i | \textbf{x}_i)} [\textbf{z}_i \textbf{z}_i^\textrm{T}] = \sigma^2 \textbf{M}^{-1} + \mathbb{E}_{p(\textbf{z}_i | \textbf{x}_i)} [\textbf{z}_i]  \mathbb{E}_{p(\textbf{z}_i | \textbf{x}_i)} [\textbf{z}_i] ^\textrm{T} \\
\mathbb{E}_{p(\textbf{z}_j | \textbf{y}_j)} [\textbf{z}_j \textbf{z}_j^\textrm{T}] = \sigma^2 \textbf{M}^{-1} + \mathbb{E}_{p(\textbf{z}_j | \textbf{y}_j)} [\textbf{z}_j]  \mathbb{E}_{p(\textbf{z}_j | \textbf{x}_j)} [\textbf{z}_j] ^\textrm{T}
\end{gathered}
\end{equation}
where $\textbf{M} = \sigma^2 \textbf{I}_k + \textbf{S}^\textrm{T}\textbf{S}$ and 
\begin{equation}
\begin{gathered}
\mathbb{E}_{p(\textbf{t}_i | \textbf{x}_i)} [\textbf{t}_i]  = \textbf{Q}^{-1}\textbf{W}^\textrm{T}(\textbf{x}_i - \mu_x) \\
\mathbb{E}_{p(\textbf{t}_i | \textbf{x}_i)} [\textbf{t}_i \textbf{t}_i^\textrm{T}] = \sigma^2 \textbf{Q}^{-1} +\mathbb{E}_{p(\textbf{t}_i | \textbf{x}_i)} [\textbf{t}_i]  \mathbb{E}_{p(\textbf{t}_i | \textbf{x}_i)} [\textbf{t}_i] ^\textrm{T}
\end{gathered}
\end{equation}
where $\textbf{Q} = \sigma^2 \textbf{I}_t + \textbf{W}^\textrm{T}\textbf{W}$. Because of the orthogonality constraint, the expectation of the outer product of $\textbf{t}_i$ and $\textbf{z}_i$ is the outer products of the expectations.
The update equations are derived by taking the derivative of the lower bound of the likelihood with respect to each of the parameters. The update equation for the shared factor loading is
\begin{equation}
\begin{split}
\tilde{\textbf{S}} = \bigg[ \sum_{i = 1}^{n} (\textbf{x}_i - \boldsymbol{\mu}_x -\textbf{W} \mathbb{E}_{p(\textbf{t}_i | \textbf{x}_i)} [\textbf{t}_i] ) \mathbb{E}_{p(\textbf{z}_i | \textbf{x}_i)} [\textbf{z}_i] ^\textrm{T} &+ \sum_{i = 1}^{m} (\textbf{y}_i - \boldsymbol{\mu}_y) \mathbb{E}_{p(\textbf{z}_i | \textbf{y}_i)} [\textbf{z}_i] ^\textrm{T} \bigg] \\& \bigg[\sum_{i = 1}^{n} \mathbb{E}_{p(\textbf{z}_i | \textbf{x}_i)} [\textbf{z}_i\textbf{z}_i^\textrm{T} ]  +  \sum_{i = 1}^{m} \mathbb{E}_{p(\textbf{z}_i | \textbf{y}_i)} [\textbf{z}_i\textbf{z}_i^\textrm{T} ] \bigg]^{-1}.
\end{split}
\end{equation}
Following the same ideas for $\textbf{W}$
\begin{equation}
\tilde{\textbf{W}} = \bigg[ \sum_{i = 1}^{n} (\textbf{x}_i - \boldsymbol{\mu}_x + \textbf{S} \mathbb{E}_{p(\textbf{z}_i | \textbf{x}_i)} [\textbf{z}_i]) \mathbb{E}_{p(\textbf{t}_i | \textbf{x}_i)} [\textbf{t}_i]^\textrm{T}\bigg] \bigg[\sum_{i = 1}^{n} \mathbb{E}_{p(\textbf{t}_i | \textbf{x}_i)}[ \textbf{t}_i \textbf{t}_i^\textrm{T} ] \bigg]^{-1}
\end{equation}
These equations can be simplified by plugging in the definitions of the expectations, as is shown in the main text.


\section{Model variants}
Several different priors can be used to obtain a sparse target loading matrix. Here we present a discussion of the pros and cons of these modeling choices. As mentioned in the manuscript, the horseshoe prior  is useful for subset selection because it has heavy tails and an infinite spike at zero. The infinite spike at zero provides strong shrinkage while the heavy tails allow coefficients to ``escape" penalization as compared to the ARD prior. In the context of sparse cLVM, the horseshoe prior is used
\begin{equation}
\begin{gathered}
\textbf{W}_{i:}|\rho_i, \tau \sim \mathcal{N}(0, \rho_i^2\tau^2\textbf{I}_t) \\
\rho_i \sim C^+(0,1), \quad \tau \sim C^+(0, b_g)
\end{gathered}
\end{equation}
In practice to use the horseshoe prior, we suggest reparameterizing the half-Cauchy distributions as inverse gamma distributions
\begin{equation}
a \sim C^+(0, b) \Longleftrightarrow a^2 | \lambda \sim \text{IG}\Big(\frac{1}{2}, \frac{1}{\lambda}\Big), \quad \frac{1}{\lambda} \sim \text{IG}\Big(\frac{1}{2}, \frac{1}{b^2}\Big)
\end{equation}
where $\text{IG}$ is the inverse gamma distribution \cite{Wand2011}. We find that standard exponential family distributions are better able to approximate the inverse gamma distributions leading to improvements during the learning phase. The variational approximation used for the inverse gamma distribution is log-normal. In the experiments section below, the sparse cLVM using the horseshoe prior is applied to the mHealth dataset.

Finally, we note that the horseshoe prior will shrink the weight values but will not set them equal to zero, therefore a thresholding rule is required to prune the weights entirely. We suggest a pruning rule that considers the posterior distribution of the scales $\rho_i^2\tau^2$. For example, prune rows of $\textbf{W}$, when the probability that the scale is less than a sufficiently small value $\delta$ is greater than $p_0$, i.e. $p(\rho_i^2\tau^2 < \delta) > p_0$.


\section{Experiments}
In this section, we present additional results on synthetic datasets and further comparisons for the real datasets.

\subsection{Synthetic data}

\begin{figure*}
	\centering
	\includegraphics{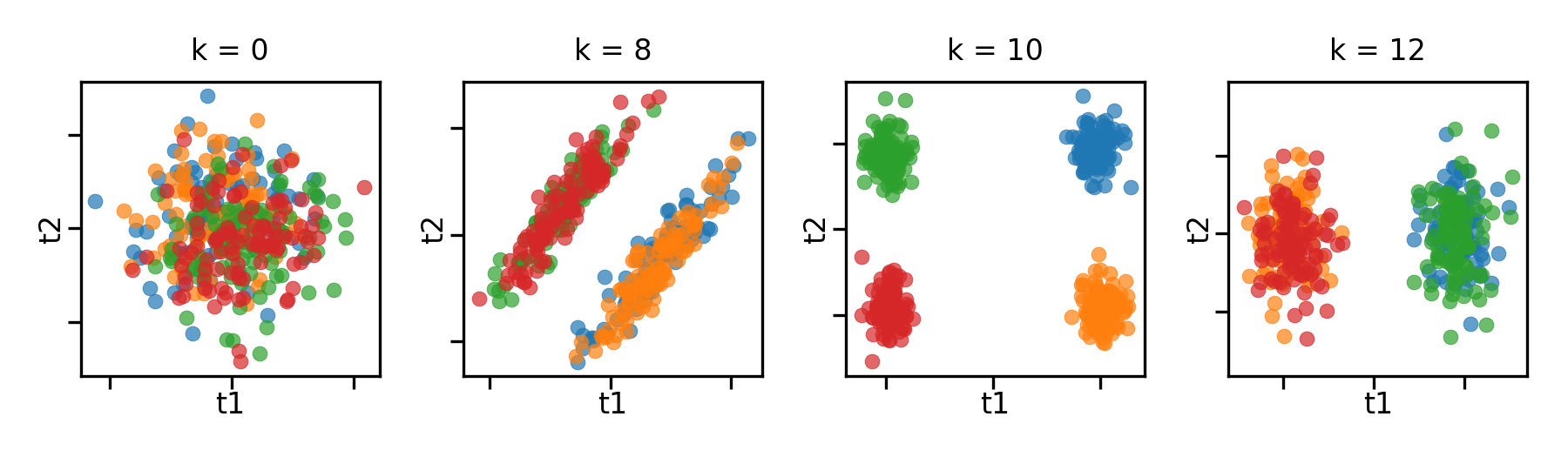}
	\caption{LVM applied to the synthetic test data. The title of the plot is the shared dimension of the latent space for the target and background samples. The dimension of the target latent space is 2. When the shared dimension is zero, the method is equivalent to applying PPCA to the target dataset.}
	\label{fig:syn}
\end{figure*}

\begin{figure*}
	\centering
	\includegraphics{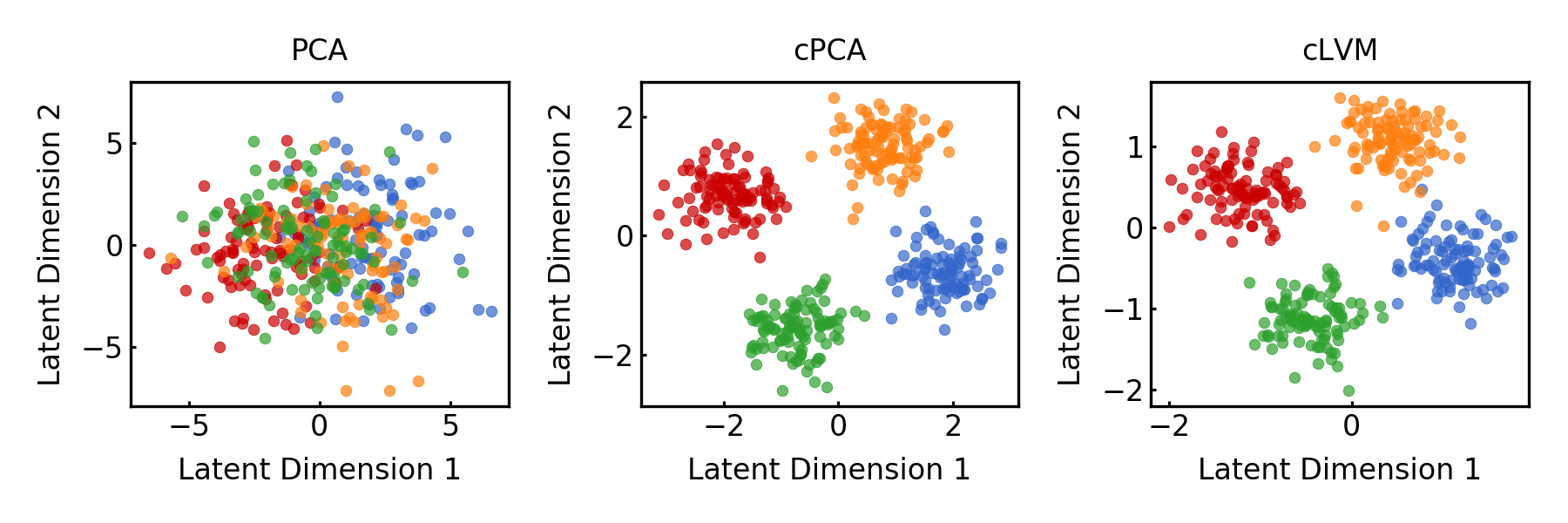}
	\caption{PCA, cPCA, and cLVM with automatic model selection applied to the generative synthetic test data. PCA is unable to discover the subtypes in the data whereas cPCA and cLVM}
	\label{fig:synGen}
\end{figure*}

\begin{figure*}
	\centering
	\includegraphics{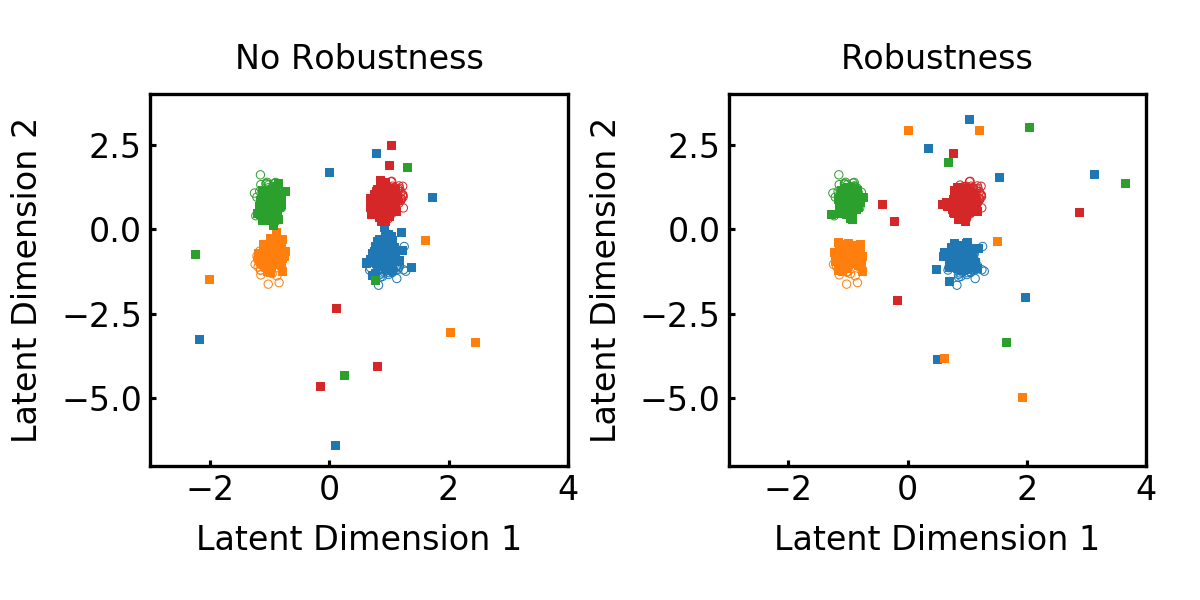}
	\caption{The target latent representation without outliers is plotted using empty circles. The left plot shows the target latent representation learned in the presence of outliers using cLVM (square points). The left plot shows the target latent representation learned in the presence of outliers using robust cLVM (square points). The robust cLVM model variant is better able to recover the target latent representation without outliers.}
	\label{fig:outlier}
\end{figure*}

\begin{table*}[]
	\centering
	\begin{tabular}{lccccc}
		& \begin{tabular}[c]{@{}c@{}}Target \\ Subgroup A\end{tabular} & \begin{tabular}[c]{@{}c@{}}Target \\ Subgroup B\end{tabular} & \begin{tabular}[c]{@{}c@{}}Target \\ Subgroup C\end{tabular} & \begin{tabular}[c]{@{}c@{}}Target \\ Subgroup D\end{tabular} & Background                       \\ \hline
		Features 1-10  & $\mathcal{N}(0,1)$  & $\mathcal{N}(0,1)$  & $\mathcal{N}(6,1)$  & $\mathcal{N}(6,1)$      & $\mathcal{N}(0,3)$ \\
		Features 11-20 & $\mathcal{N}(0,1)$                              &$\mathcal{N}(3,1)$                              & $\mathcal{N}(0,1)$                              & $\mathcal{N}(3,1) $                             & $\mathcal{N}(0,1)$  \\
		Features 21-30 & $\mathcal{N}(0,10)$                             & $\mathcal{N}(0,10) $                            & $\mathcal{N}(0,10)$                             & $\mathcal{N}(0,10)$                             & $\mathcal{N}(0,10)$
	\end{tabular}
	\caption{A summary of the distributional assumptions used to generate the synthetic data used in analysis.}
	\label{tab:dist}
\end{table*}

cLVM is demonstrated on two synthetic datasets. The first is based on the synthetic dataset proposed in \cite{Abid2018}. 30-dimensional data is simulated using the distributional assumptions summarized in Table~\ref{tab:dist}. The target and background datasets each have 400 samples. The target latent representation for different choices of the shared latent dimensionality are shown in Fig.~\ref{fig:syn}. The second synthetic dataset is generated using the distributional assumptions using eqn 3 where the target latent variables have cluster specific means. The target latent representation for different dimensionality reduction techniques are shown in Fig.~\ref{fig:synGen}. cLVM with automatic model selection is able to correctly discover the dimensionality of the shared factor loading. A third example is shown by adding outliers to the second synthetic dataset. Outliers are drawn from a uniform distribution [-20, 20] and 20 outliers each are added to the target and background datasets. cLVM and robust cLVM are applied to the dataset. The target latent representations are shown in Fig.~\ref{fig:outlier}.

\subsection{Mice Protein Expression Data}
The latent representation of the mice protein expression data is also compared to PCA (see Fig.~\ref{fig:micePCCA}, first row). For all levels of missing data, PCA is unable to recover the latent subgroups.
\begin{figure*}
	\centering
	\includegraphics[width=\textwidth]{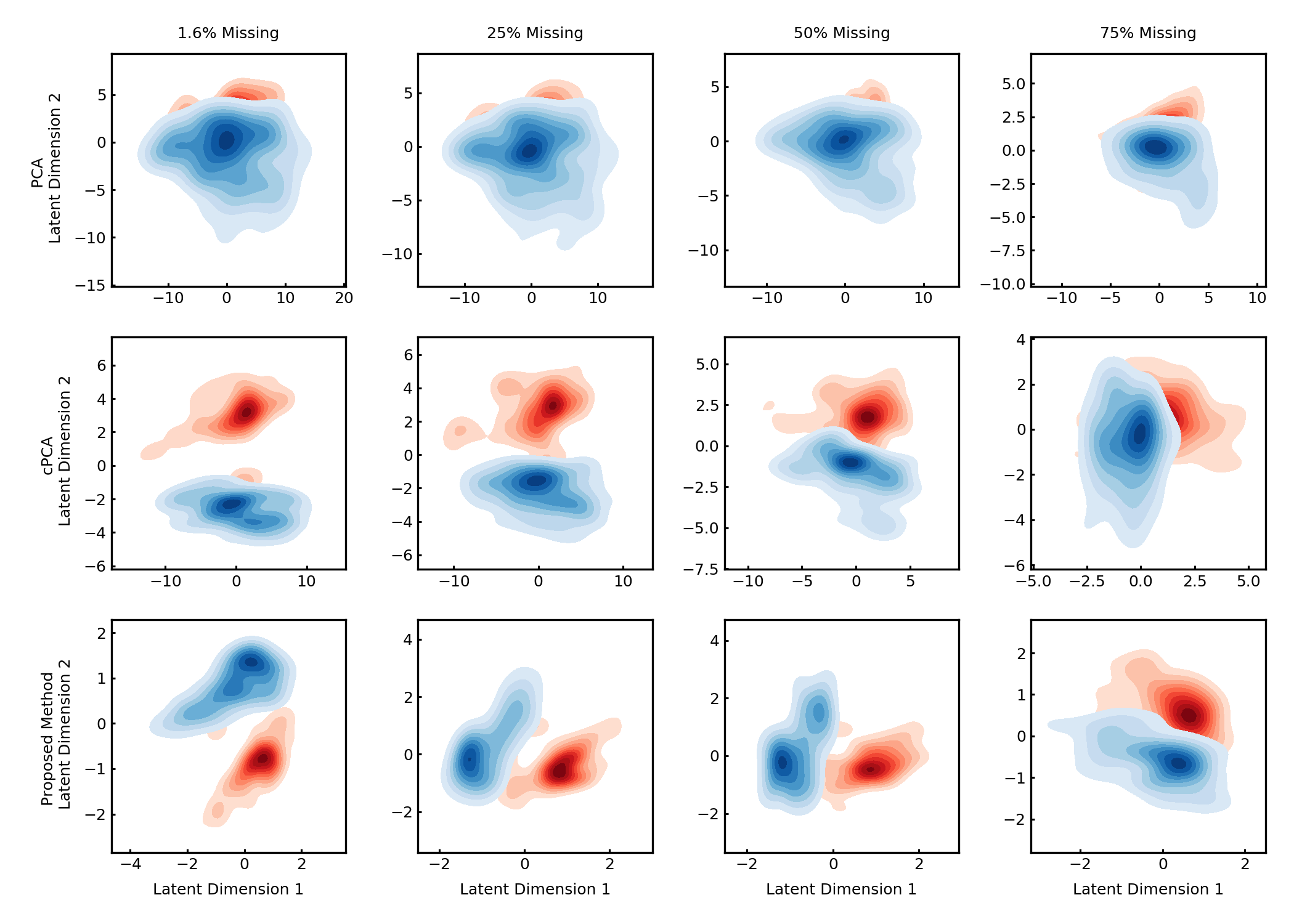}
	\caption{Density plots of the subgroups revealed in the target latent representation of the mice protein expression data. Red and blue points are the control and trisomic mouse samples, respectively. The rows use PCA, cPCA, and robust cLVM to learn the latent representation, respectively. Each column uses a different level of missing data, starting with the leftmost column containing the natural level of missing data. PCA is unable to perform subgroup discovery and LVM is better able to perform subgroup discovery in the presence of missing data.}
	\label{fig:micePCCA}
\end{figure*}

\subsection{Single Cell RNA-Seq Data}
The latent representation of the single cell RNA-Seq data is compared to PCA and cPCA in Fig.~\ref{fig:RNASeq}. PCA is unable to recover the latent subgroups. cPCA and cLVM recover similar latent representation, however cLVM does not require a priori selection of tuning parameters whereas cPCA uses a heuristic to select the value of $\alpha$, which controls the tradeoff between maximizing the variance in the target dataset and minimizing the variance in the background dataset.
\begin{figure*}
	\centering
	\includegraphics[width=\textwidth]{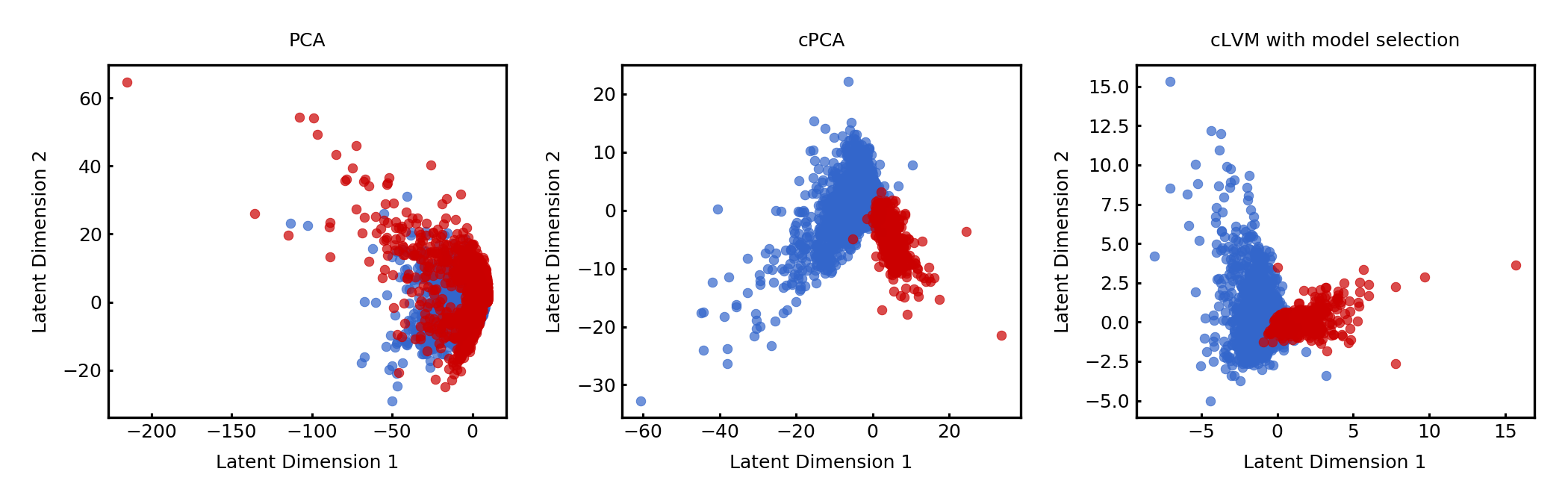}
	\caption{The latent representation of the single cell RNA-Seq data. Red and blue points are the post- and pre-treatment samples, respectively. From left to right, the plots use PCA, cPCA, and cLVM with model selection to learn the latent representation, respectively. PCA is unable to perform subgroup discovery. cPCA is able to perform subgroup discovery and uses a heuristic to selection $\alpha$, the tuning parameter for the empirical covariance matrix (eqn.~1). cLVM with model selection finds a similar latent representation to cPCA without the need to select the shared dimensionality a priori by using automatic relevance detection.}
	\label{fig:RNASeq}
\end{figure*}

\subsection{mHealth Sensor Data}
\begin{figure}
	\centering
	\includegraphics[width=\textwidth]{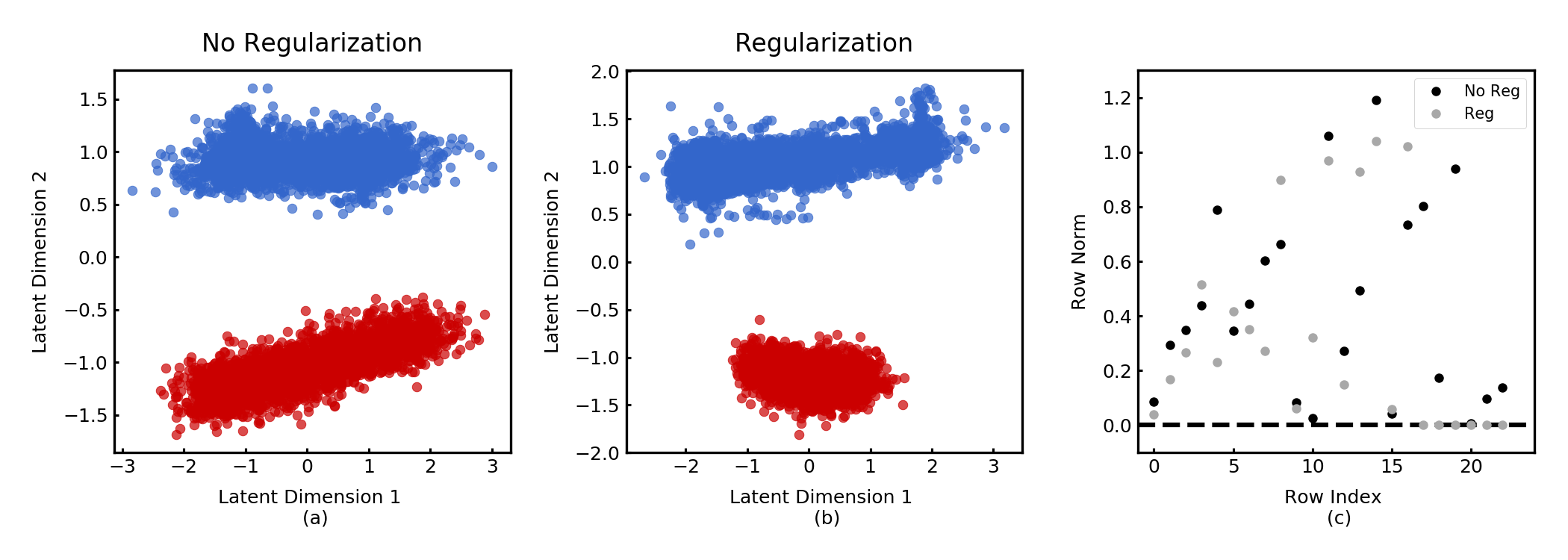}
	\caption{The target latent representations for the mHealth dataset with and without regularization. In both cases, there are clearly two subgroups, which correspond to the squatting and cycling activities. (c) The row norms of the target factor loading matrix. In the case with regularization, the measurements corresponding the to the magnetometer (last six measurements) are zero.}
	\label{fig:mHealth}
\end{figure}

\begin{figure}
	\centering
	\includegraphics[width=0.67\textwidth]{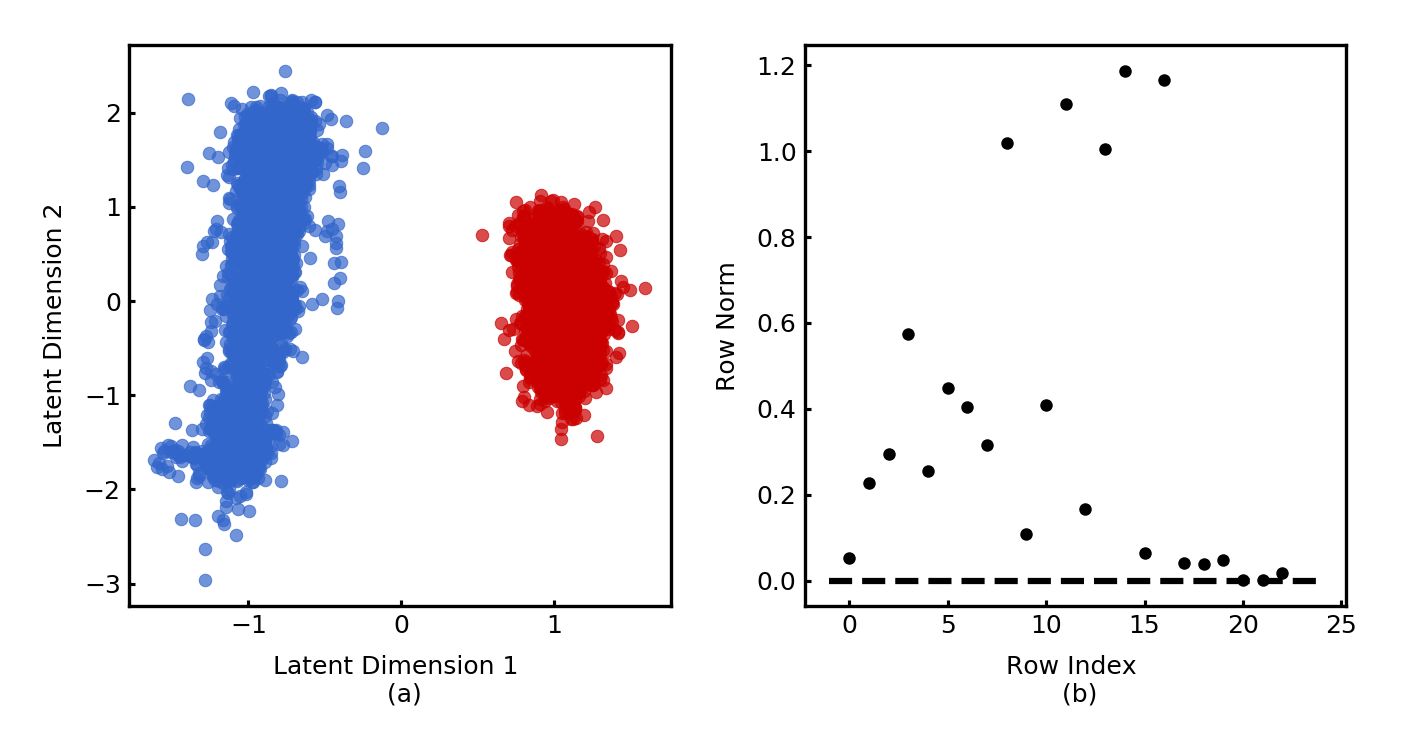}
	\caption{(a) The target latent representation for the mHealth dataset using a horseshoe prior for the target factor loading. (b) The row norms of the target factor loading. The sensor measurements are ordered the same as in Fig.~\ref{fig:mHealth}.}
	\label{fig:mHealthHS}
\end{figure}

\begin{figure}
	\centering
	\includegraphics[width=\textwidth]{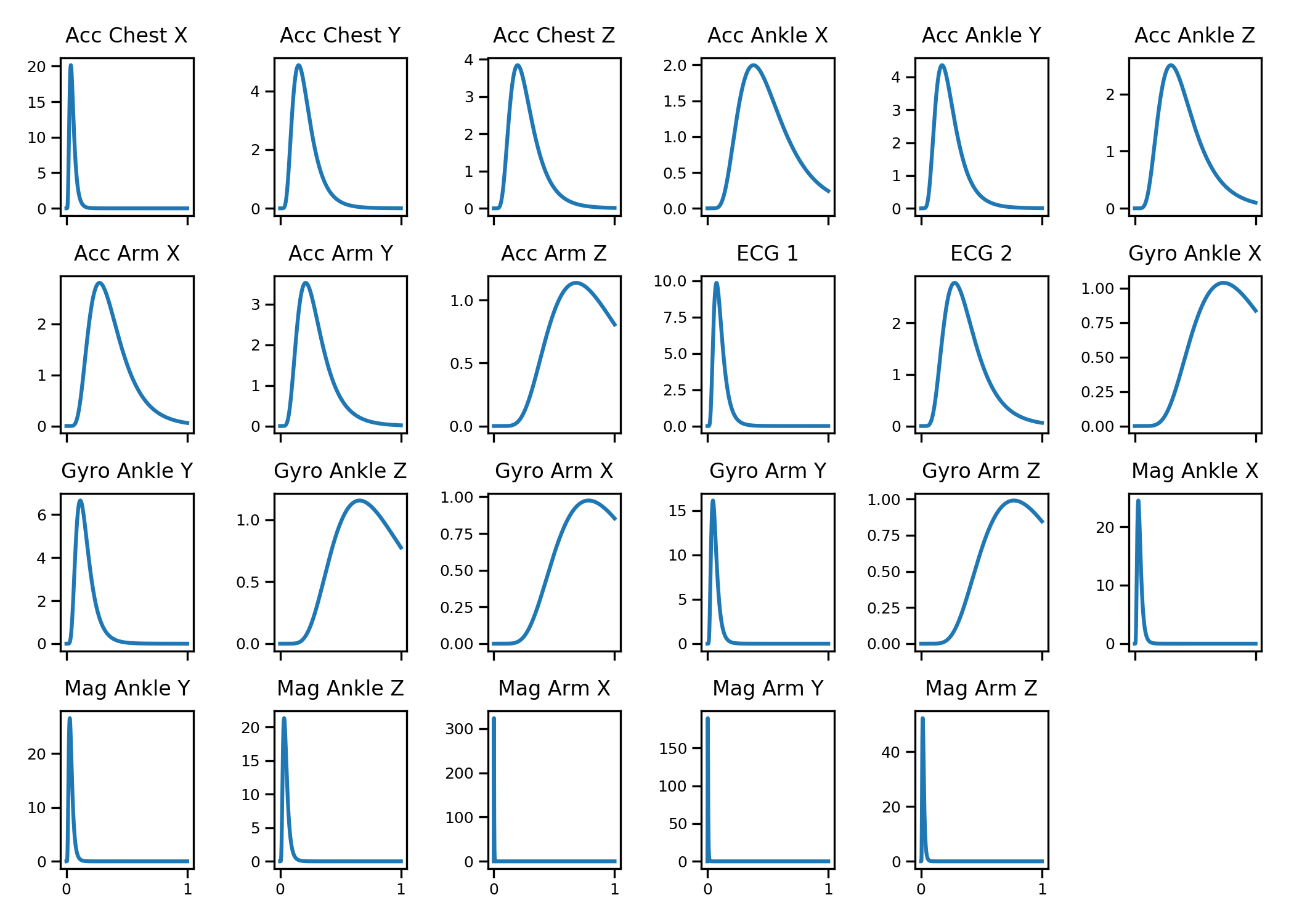}
	\caption{The distribution of the scales ($\rho_i^2\tau^2$) for each of the rows of the target factor loading matrix $\textbf{W}$. The title of the plot is the type of sensor: accelerometer (Acc), electrocardiogram (EGC), gyroscope (Gyro), and magnetometer (Mag). X, Y, and Z indicate the axis of the measurement. Note that the y-axis is not shared amongst the plots and the magnetometer measurements are highly peaked near zero. Increasing values of a pruning threshold will cause an increasing number of the magnetometer sensors coefficients to go to zero. A reasonably chosen threshold may also result in pruning the Acc Chest X-axis sensor.}
	\label{fig:mHealthHS_dist}
\end{figure}

The target latent representation for the mHealth data with and without regularization is presented in Fig.~\ref{fig:mHealth}. In both cases, there are two clear subgroups in the target latent representation, which correspond to squatting and cycling. For comparison to a probabilistic result, we also present the target latent representation using the horseshoe prior in Fig.~\ref{fig:mHealthHS}. By using the horseshoe prior, we also have information about the distribution of the factor loading matrix. The distributions of the scales which govern the variance of the factor loading matrix are shown in Fig.~\ref{fig:mHealthHS_dist}

\subsection{MNIST Digits on Grass}
\begin{figure*}
	\centering
	\includegraphics[width=\textwidth]{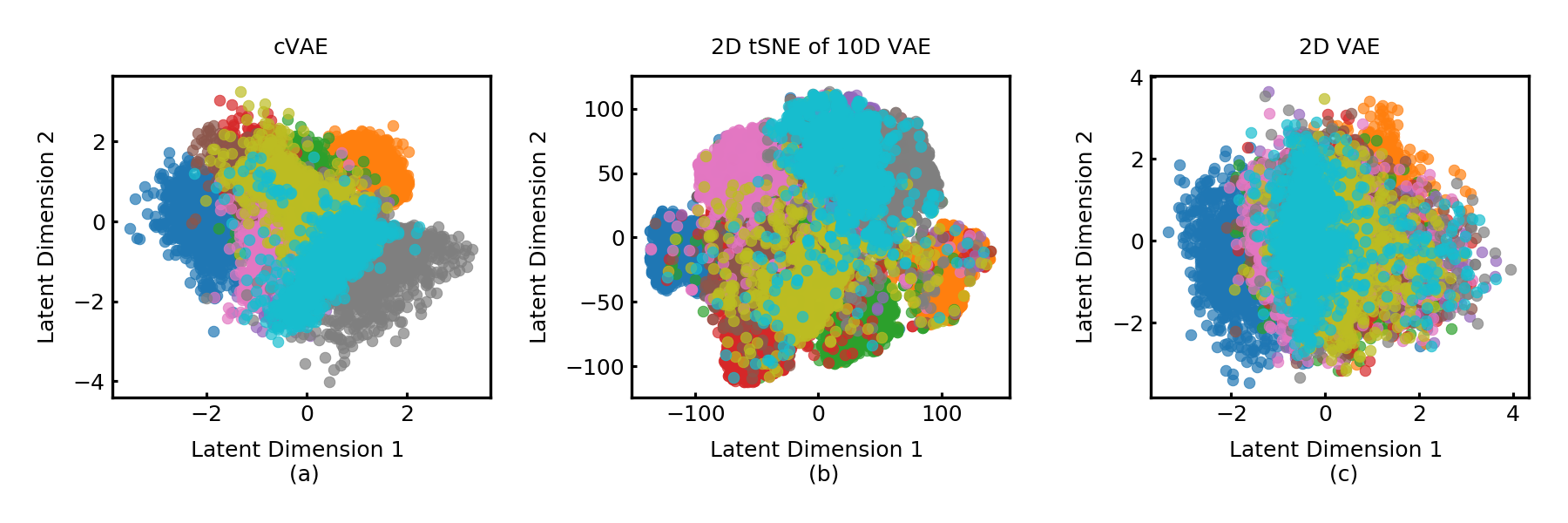}
	\caption{Latent projections from cVAE, VAE with 2D latent space, 2D tSNE projections of VAE with 10 dimensional latent space.}
	
	\label{fig:MNIST}
\end{figure*}

Latent representations recovered by contrastive and regular VAEs on the MNIST on Grass dataset are shown in Fig.~\ref{fig:MNIST}.

\end{document}